# Myocardial Region-guided Feature Aggregation Net for Automatic Coronary artery Segmentation and Stenosis Assessment using Coronary Computed Tomography Angiography

Ni Yao[1], Xiangyu Liu[1], Danyang Sun[1], Chuang Han[1], Yanting Li[1], Jiaofen Nan[1], Chengyang Li[2], Fubao Zhu[1], Weihua Zhou[3,4*], Chen Zhao[5*]

[1]School of Computer Science and Technology, Zhengzhou University of Light Industry, Zhengzhou 450001, Henan, China
[2]School of Information Management and Engineering, Shanghai University of Finance and Economics, China
[3]Department of Applied Computing, Michigan Technological University, Houghton, MI, USA
[4]Center for Biocomputing and Digital Health, Institute of Computing and Cybersystems, and Health Research Institute, Michigan Technological University, Houghton, MI, USA
[5]Department of Computer Science, Kennesaw State University Marietta, GA, USA

*Correspondence:
Weihua Zhou, Ph.D.
Assistant Professor
Department of Applied Computing
Michigan Technological University, Houghton, MI, USA
Email: whzhou@mtu.edu

Chen Zhao, Ph.D.
Assistant Professor
Department of Computer Science
Kennesaw State University Marietta, GA, USA
Email: czhao4@kennesaw.edu

**Abstract**

Coronary artery disease (CAD) remains a leading cause of mortality worldwide, requiring accurate segmentation and stenosis detection using Coronary Computed Tomography angiography (CCTA). Existing methods struggle with challenges such as low contrast, morphological variability and small vessel segmentation. To address these limitations, we propose the Myocardial Region-guided Feature Aggregation Net, a novel U-shaped dual-encoder architecture that integrates anatomical prior knowledge to enhance robustness in coronary artery segmentation. Our framework incorporates three key innovations: (1) a Myocardial Region-guided Module that directs attention to coronary regions via myocardial contour expansion and multi-scale feature fusion, (2) a Residual Feature Extraction Encoding Module that combines parallel spatial channel attention with residual blocks to enhance local-global feature discrimination, and (3) a Multi-scale Feature Fusion Module for adaptive aggregation of hierarchical vascular features. Additionally, Monte Carlo dropout f quantifies prediction uncertainty, supporting clinical interpretability. For stenosis detection, a morphology-based centerline extraction algorithm separates the vascular tree into anatomical branches, enabling cross-sectional area quantification and stenosis grading. The superiority of MGFA-Net was demonstrated by achieving an Dice score of 85.04%, an accuracy of 84.24%, an HD95 of 6.1294 mm, and an improvement of 5.46% in true positive rate for stenosis detection compared to3D U-Net. The integrated segmentation-to-stenosis pipeline provides automated, clinically interpretable CAD assessment, bridging deep learning with anatomical prior knowledge for precision medicine. Our code is publicly available at http://github.com/chenzhao2023/MGFA_CCTA.

**Highlights**

- The segmentation process is guided by myocardial contour expansion, focusing attention on coronary anatomy to improve segmentation accuracy.
- Spatial-channel attention mechanisms integrate myocardial region features, enhancing vascular contrast and feature discrimination.
- Adaptive multi-scale feature fusion dynamically aggregates hierarchical coronary details to capture fine anatomical structures.
- Stenosis detection for coronary artery contouring

# 1 Introduction

Coronary artery disease (CAD) is the major cause of mortality worldwide, with an annual mortality rate that exceeds one-fourth of all deaths. Furthermore, the incidence and mortality rates of CAD continue to rise [1]. The pathological hallmark of CAD is the accumulation of cholesterol plaques on the inner walls of the coronary arteries, leading to arterial stenosis and subsequent myocardial ischemia [2]. As the ischemia worsens, patients may experience symptoms such as chest pain (angina pectoris) and shortness of breath. Furthermore, the instability of the stenotic areas can result in acute coronary occlusion, which in turn leads to myocardial infarction [3, 4].

Accurate diagnosis and early risk assessment of CAD are crucial for effective patient management. Coronary Computed Tomography angiography (CCTA), a non-invasive and widely utilized imaging technique, provides images of a quality that is comparable to that of traditional Coronary Angiography (ICA) [5]. However, the localization and measurement of the severity of coronary artery stenosis typically requires manual assessment by specialized cardiologists, which is not only time-consuming but also prone to misdiagnosis or missed diagnoses [6]. The development of high-quality, fully automated coronary artery segmentation techniques has therefore become a priority. However, the application of automatic segmentation techniques using CCTA faces multiple challenges [7], such as the coronary artery system is composed of branches of various sizes, and its morphology and location vary significantly among individuals. In addition, the coronary arteries exhibit structural and imaging feature similarities with other vessels, which can cause confusion and result in incorrect identification during the segmentation process [8].

In this paper, we propose a Myocardial Region-guided Feature Aggregation Net (MRFA-Net) based on a myocardial region-guided U-shaped network. This model demonstrates detail-awareness, enabling effective discrimination of small vessel segments, and performs multi-scale fusion of decoded features to enhance the segmentation accuracy. The MRFA-Net consists a dual encoder and a single decoder. In the Myocardial Region-guided Module (MRG), images from the segmented and processed myocardial regions are utilized, and feature fusion is performed by the image encoder at each down sampling layer. Regional Features from the MRG are fused in the image encoder using the Residual Feature Extraction Encoding Module (RFEE) to enhance the perception of coronary details in the myocardial region. Additionally, a Multi-scale Feature Fusion Module (MSFF) is designed to integrate rich features from different scales in the decoder, further boosting the segmentation accuracy. The synergistic effect of the MRG, RFEE, and MSFF modules, in conjunction with Monte Carlo Dropout, leads to an impressive coronary artery segmentation performance. The proposed MRFA-Net achieved a Dice Similarity Coefficient of 85.04%, precision of 84.24%, recall of 87.27%, and a 95% Hausdorff Distance (HD95) of 6.1294 mm.

The results from MGFA-Net demonstrate the centerlines of the coronary arteries, which are then used to divide the arteries into multiple segments. For each segment, the cross-sectional area is measured at each location and the degree of stenosis is scored according to the corresponding criteria. Compared to the results from 3D U-Net, our method achieved an increase in True Positive Rate (TPR) by 5.46%, respectively, and a decrease in Absolute Root Mean Square Error (ARMSE) and Relative Root Mean Square Error (RRMSE) by 0.0037 and

0.0201, respectively

The main contributions of this work are summarized as follows:

1. A novel network guided by myocardial region is proposed for 3D coronary artery segmentation using CCTA, leveraging myocardial contour expansion to direct computational attention toward coronary anatomy, thereby improving delineation accuracy in regions with low contrast.

2. A hierarchical encoder integrates five RFEE modules to fuse myocardial region features during downsampling, enhancing local-global feature discrimination while preserving vascular morphology.

3. The decoder employs adaptive MSFF to dynamically aggregate hierarchical vessel features across scales, retaining fine structural details and boosting segmentation precision.

4. A morphology-based algorithm extracts coronary centerlines from segmented vessels, enabling branch-specific cross-sectional area quantification and systematic stenosis severity grading for clinical interpretability.

## 2 Related Work

In recent years, deep learning has achieved significant progress in the field of medical image segmentation. Specifically, CNN-based deep learning has exhibited both remarkable precision and automation levels in the domain of coronary artery segmentation. In the context of 3D coronary CT angiography (CCTA) data, 3D-CNNs have been the prevailing approach [9]. These networks are able to process 3D imaging data directly, thereby leveraging the inherent spatial information to enhance the accuracy of segmentation. By capturing the overall structure of coronary arteries in 3D space, 3D-CNNs produce more complete and accurate segmentation results. Multi-view fusion methods, which integrate information from multiple perspectives, have been demonstrated to enhance the recognition of coronary arteries out of the limitations of single-view approaches. The U-Net [10] and V-Net [11] are representative models that effectively retain spatial resolution through multi-scale feature extraction and skip connections, significantly improving segmentation performance. However, it should be noted that, at present, due to the attention mechanism and the introduction of Transformer, basic CNNs are no longer able to meet the demand for coronary artery segmentation accuracy.

In order to achieve enhanced segmentation of coronary arteries, Yu et al. [12] proposed a densely connected convolutional neural network, DenseVoxNet, for the automatic segmentation of vascular structures from 3D MRI (Magnetic Resonance Imaging). Mou et al. [13] introduced $CS^2$-Net, a segmentation network that utilises a dual-channel attention mechanism to capture the complex representations of curvilinear structures. Song et al. [14] employed dense blocks and residual blocks to extract representative features for coronary artery segmentation. Xia et al. [15] proposed ER-Net, a segmentation network that preserves spatial edge information to improve segmentation performance. Duan et al. [16] designed ECA-UNet, which achieves cross-channel interaction and effectively enhances the segmentation performance of coronary arteries. Furthermore, transfer learning-based methods combined with the U-Net architecture have demonstrated excellent performance in CT coronary angiography images, especially in segmenting small vessel segments and low-contrast images [17]. Geometric cascade neural network-based methods for coronary artery segmentation and vessel vectorization have achieved high-precision results by combining geometric feature extraction with multi-level networks[18].Vision Transformer [19] has shown significant performance improvements in this context. DiffCAS [20] is a network based on the Denoising Diffusion Probabilistic Model (DDPM), which uses Swin Transformer [21] to extract semantic information from CCTA images during the denoising diffusion process, thereby improving segmentation performance. The majority of these methods employ transformers or attention mechanisms to extract image features. In contrast to these approaches, our method employs myocardial regions as a guide, integrating attention mechanisms to extract image features. This approach enhances the model's sensitivity to the target area, leading to an improvement in segmentation performance.

**3 Method**

The present study is divided into two sections, (1) automatic coronary artery segmentation and (2) stenosis detection based on artery segmentation. Following the acquisition of coronary artery segmentation results, stenosis detection was conducted for each arterial segment contour, thereby facilitating a more comprehensive evaluation of the severity of stenosis in each arterial segment.

**3.1 Coronary artery segmentation**

**3.1.1 Network architecture**

The overall framework of MGFA-Net is shown in Figure 2. The MGFA-Net framework is designed to address the challenges associated with coronary artery segmentation, including low contrast, blurred boundaries, irregular shapes, and thin branches (see Figure 1). To address these challenges, the MGFA-Net framework incorporates three modules: MRG, RFEE, and MSFF.

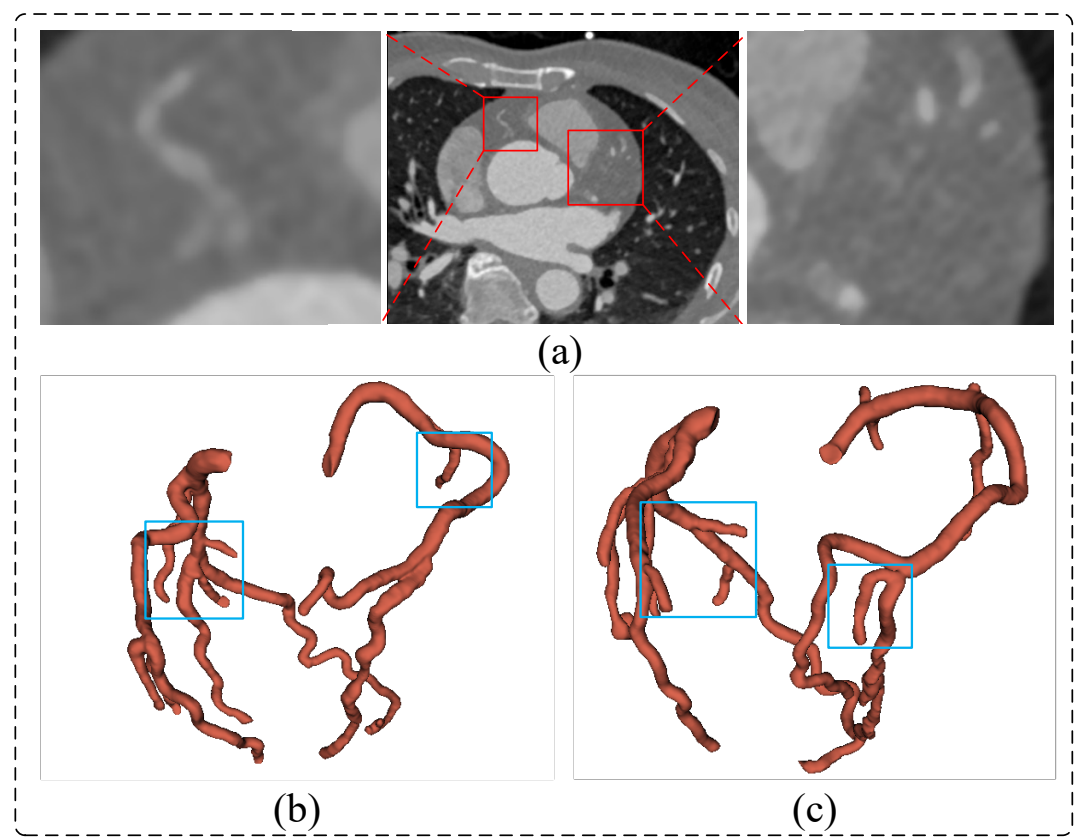

Figure 1 Morphology of coronary arteries on CCTA and visualization in 3D (a) Low contrast and blurred boundaries of coronary arteries, (b) and (c) Two examples of the small branches

of coronary arteries.

The proposed MGFA-Net is based on the U-shaped architecture with an additional encoder, namely the Myocardial Region-guided Encoder (MRG). The MRG consists of two parts. The first part involves the segmentation of the myocardium and a series of operations, including the expansion of the myocardial region. The second part works in conjunction with the image encoder to progressively fuse myocardial region features with image features during the downsampling process. This underscores the significance of myocardial region features and mitigates the impact of noise from other regions during the coronary artery segmentation process. The proposed RFEE module enhances feature extraction using both the channel and spatial attention mechanisms. During the decoding process, the MSFF is introduced to integrate multi-scale image information, thereby enhancing the performance of coronary artery segmentation.

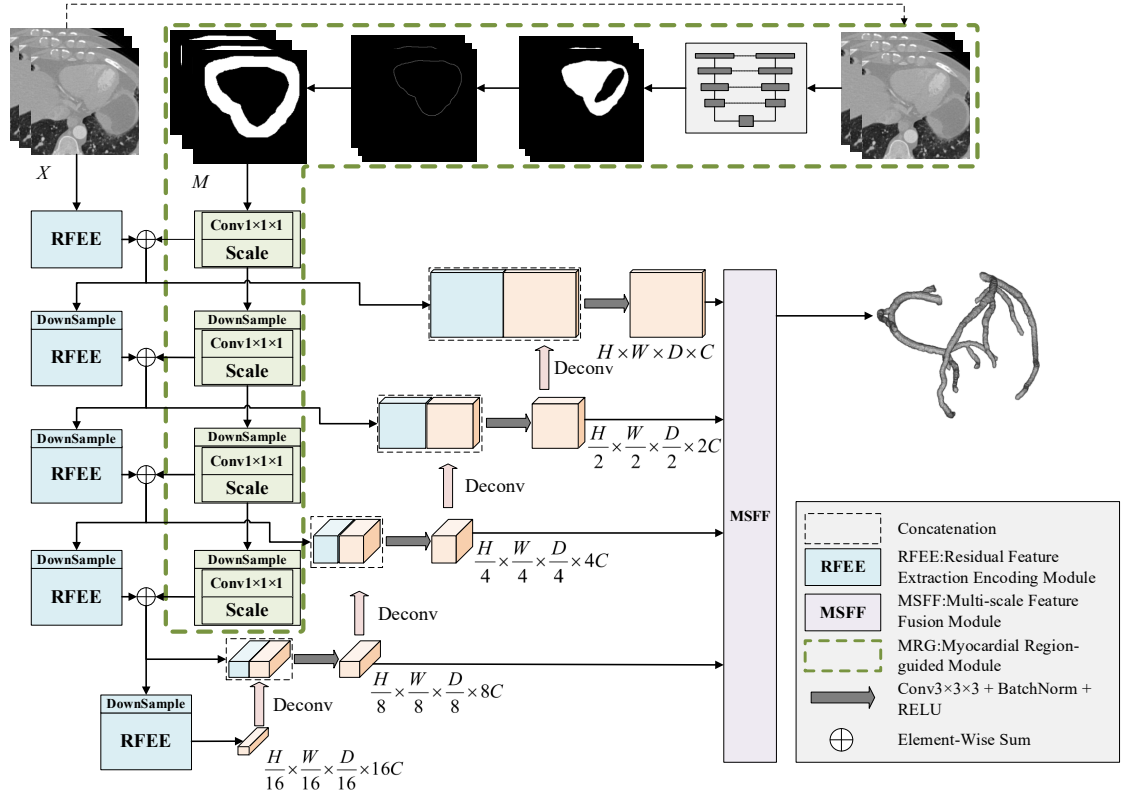

Figure 2 Architecture of the proposed MGFA-Net

### 3.1.2 Myocardial Region-guided Encoder

From a clinical perspective, coronary arteries originate from the aorta and supply blood to the myocardium [22-24]. However, the segmentation of coronary arteries within the context of CCTA is influenced by surrounding organs or tissues, leading to over-segmentation or under-segmentation. To address this issue, the MRE is developed, integrating prior knowledge, i.e. the location of the myocardium, into the coronary artery segmentation process. This enables the model to focus more on relevant information within the myocardial region, thereby improving segmentation performance.

The initial step involves using a trained myocardial segmentation model to delineate the myocardial regions. In detail, we trained a myocardial segmentation model based on U-Net [10], which includes an internal dataset containing 60 CCTA images and manually annotated myocardial regions. We divided 45 images for training and 15 images for validation. On the validation set, it achieved 88.93% in dice coefficient similarity, 90.19% in precision, and 88.26% in recall. This model is used for extracting myocardium for the rest CCTAs in the entire data to guide coronary artery segmentation.

Given the observation that the myocardium manifests as a contiguous region within the image, post-processing is performed by identifying and retaining the largest connected component. Consequently, myocardial contour extraction and expansion operations are conducted on the cross-sections of the image, which indicates the 2D slices of the coronary arteries and myocardium at specific points along the length of the arteries, as illustrated in Figure 3.

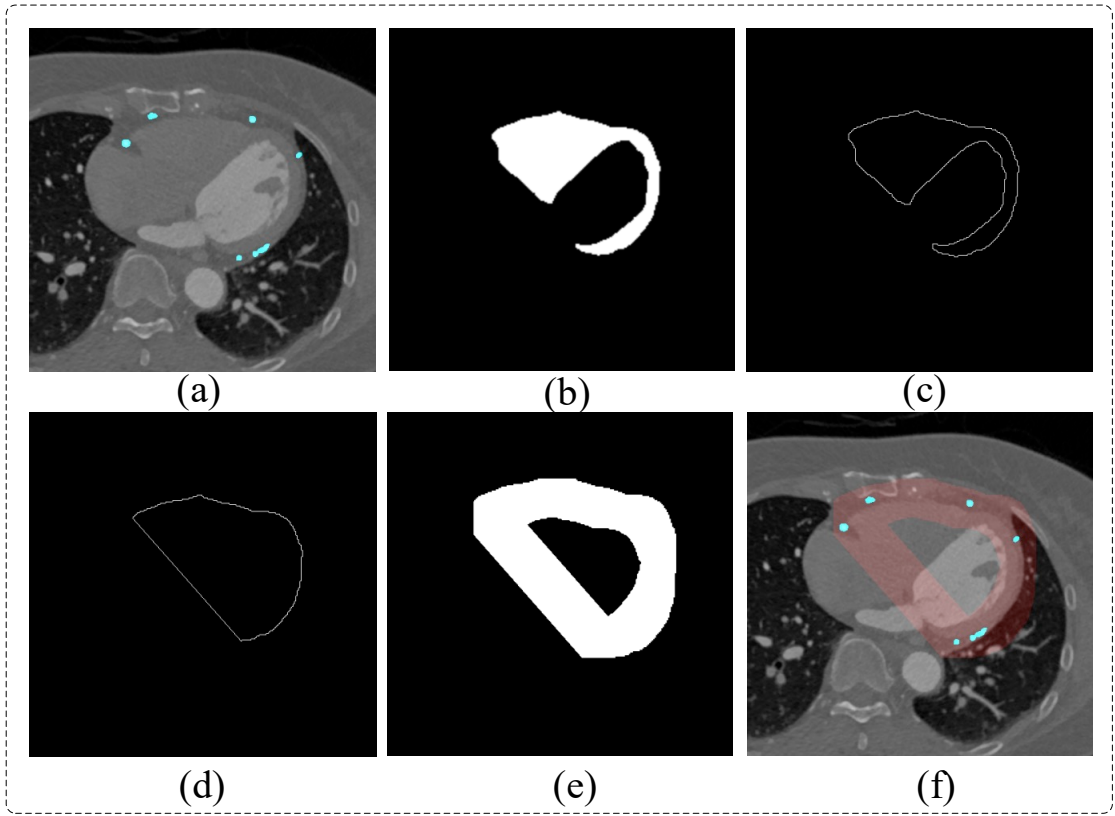

Figure 3. Acquisition of myocardial extension regions (a) slides of the CCTA image, (b) binary map of the myocardial segmentation results, (c) the contour of the myocardium, (d) Closed myocardial contour, (e) Myocardial contour expansion binary map, (f) Visualization of myocardial expansion areas on CCTA images.

In the initial phases in Figure 3 (a) and (b), the myocardial region is extracted and post-processed using the trained U-Net. In steps (b) and (c), the epicardial contour is extracted from the myocardium. To cover the entire heart region, the myocardial contour is then smoothed in steps (c) and (d). The region is expanded in steps (d) and (e), using dilation operation, with a dilation kernel size of 50 and iterations set at 1. Figure 3 (f) visualizes the expanded region, where the cyan colour represents the coronary artery region and the transparent red represents the expanded myocardial region.

As illustrated by the green dashed line in Figure 2, the myocardial region-guided encoder (MRGE) continuously fuses the expanded myocardial region obtained from the first part with image features at each downsampling step. This operation ensures that the region is better highlighted during the segmentation process, enabling the model to focus more on features within the region and reducing the impact of background noise. Assuming the input $M$ is the binary map of the expanded myocardial region with size $1 \times H \times W \times D$, where 1 represents the number of feature channels, and $H$, $W$, and $D$ represent the dimensions of the input features, the Myocardial Region-guided Module (MRG) is represented in Eq. 1:

$$M_i = \text{Scale} \times \text{DownSample}(f^{1\times1\times1}(M_{i-1})) \qquad (1)$$

where $M_{i-1}$ represents the input feature map of the expanded myocardial region at layer $i - 1 (i = 1,2,3,4)$, and $M_i$ represents the output feature map of the expanded myocardial region at layer i. $Scale$ denotes a trainable parameter with an initial value set to 0.1. $DownSample$ indicates downsampling using max pooling, and $f^{1\times1\times1}$ represents a convolution operation with a kernel size of $1 \times 1 \times 1$, which is used separately in each layer. $C$ is set as 16 in Figure 2. Through these operations across four layers, myocardial region features at different scales are obtained, denoted as $M_1$, $M_2$, $M_3$, and $M_4$, respectively.

**3.1.3 Residual Feature Extraction Encoder (RFEE)**

The accurate segmentation of coronary arteries is predicated on the extraction of complex anatomical structures and the identification of collateral arteries. In CNNs, the varying sizes of arteries result in different manifestations in feature maps at various scales. As the network depth increases, the model is able to capture broader and richer high-level semantic features, which are crucial for identifying regions with significant semantics in high-level features. Conversely, low-level features encapsulate abundant spatial information, thereby facilitating the precise reconstruction of details within high-level features. The effective integration of high-level and low-level features has been shown to enhance the model's ability to segment the complex anatomical structures of coronary arteries, thereby improving the accuracy of vessel prediction. The RFEE module, designed to address these challenges, comprises three distinct components: the residual block, the spatial attention module, and the channel attention module, as shown in Figure 4.

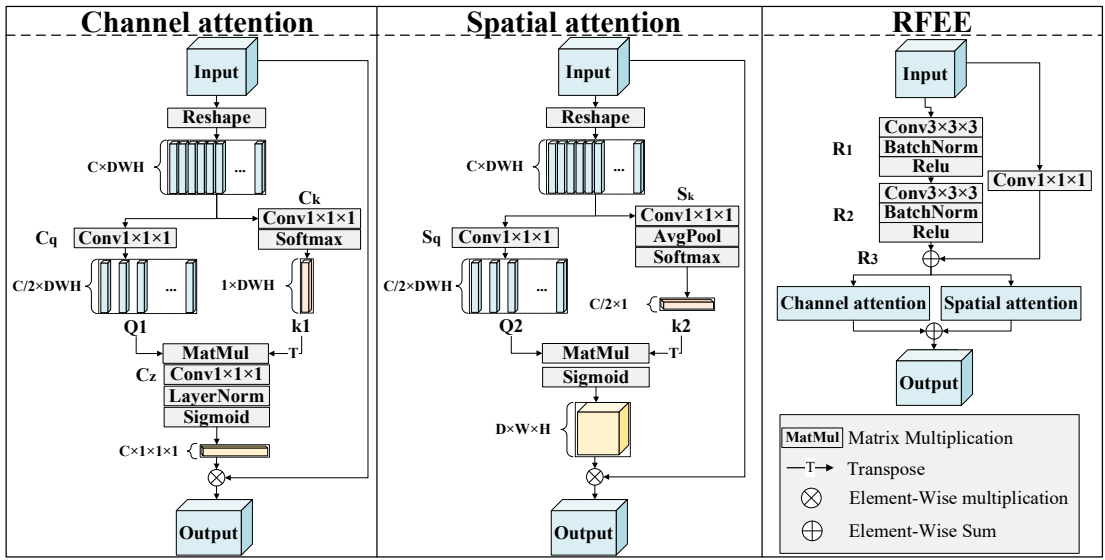

Figure 4. The architecture of the Residual Feature Extraction Encoder

Segmenting coronary arteries is challenging due to significant variations in their morphology, size, and relationships with surrounding structures across different patients [25]. Furthermore, coronary artery images frequently exhibit low contrast, and the surrounding structures display analogous imaging characteristics, which can readily result in confusion and errors during the segmentation process[14]. We employ residual blocks (RBs) to extract image features. RBs have become a popular approach in deep learning models, which extract deep-level features while mitigating the vanishing gradient problem in deep networks.[26]. We additionally employ an attention mechanism to extract more discriminative features. Conventional serial attention mechanisms process spatial and channel attention sequentially [27, 28]. However, this approach fails to fully leverage the respective advantages of spatial and channel attention, leading to redundant or lost information transfer [29, 30]. To address this issue, the RFEE module employs a parallel processing mechanism for channel and spatial attention, capturing attention weights for channels and spatial positions separately through self-attention mechanisms to fully utilise the modelling capabilities of the self-attention structure [31]. The difference between parallel and serial spatial and channel attention will be described in the experimental section. The detailed structures of spatial and channel attention are shown in Figure 4.

**Residual block:** Specifically, assuming the input $X \in \mathbb{R}^{1 \times H \times W \times D}$ represents the CCTA grayscale image, and $X_i (i = 1,2,3,4)$ denotes the feature maps of each layer, after each input feature $X_i$, the feature first passes through a residual block containing two convolutional blocks. Each convolution block includes a convolutional layer with the kernel size of $3 \times 3 \times 3$, a batch normalization, and a ReLU activation function. Finally, a residual operation is applied to obtain the feature $R_i$. This process can be summarized as follows:

$$R_1 = \text{ReLU}\left(\text{BN}\left(f^{3\times3\times3}(X_i)\right)\right) \tag{2}$$

$$R_2 = \text{ReLU}\left(\text{BN}(f^{3\times3\times3}(R_1))\right) \tag{3}$$

$$R_3 = R_2 + \text{Conv}_{1\times1\times1}(F_i) \tag{4}$$

where, $f^{3\times3\times3}$ denotes a convolutional operation with a kernel size of $3 \times 3 \times 3$, which is used separately in each layer. BN represents batch normalization, and $\text{Conv}_{1\times1\times1}$ represent convolutional layers with kernel sizes of $1 \times 1 \times 1$ which is used separately in each layer. This process generates features $R_1, R_2, R_3$ layer by layer, with each layer incorporating the operation of the residual block.

**Channel attention:** Initially, to capture cross-channel spatial importance while enabling subsequent reintroduction of channel-specific details, two convolutional layers with the kernel size of $1 \times 1 \times 1$ are applied to reduce the input channels to $C/2$ and 1, respectively. The feature map, which has a channel size of 1, then passes through a SoftMax function. The application of the SoftMax function to the smaller feature map enables the allocation of elevated attention weights to salient features. The resulting feature map is then multiplied with the feature map of channel size $C/2$ to obtain the attention map. Finally, a $1 \times 1 \times 1$ convolutional operation is employed to restore the attention map to its original input size. Layer normalisation and a Sigmoid activation function are then applied to dynamically map the attention map, which helps to recover some lost details by mapping the originally compressed attention map to an appropriate range. Finally, the input is multiplied with the attention map to obtain the feature matrix with enhanced channel features. The aforementioned process can be represented as follows:

$$R_3^{\text{channel}} = R_3 \otimes \sigma \left( LN \left( C_v \left( C_q(R_3) \times \text{SoftMax}(C_k(R_3)) \right) \right) \right) \tag{5}$$

where LN denotes Layer Normalization, $C_q$, $C_k$, and $C_v$ represent convolutional layers with kernel sizes of $1 \times 1 \times 1$, respectively. $\sigma$ is the Sigmoid activation function, and $\times$ indicates matrix multiplication, $\otimes$ represents element-wise multiplication.

**Spatial Attention:** In the spatial attention module, the self-attention mechanism is also employed to extract the attention map. Two $1 \times 1 \times 1$ convolutional operations are used to reduce the input channels to $C/2$, respectively. In contrast to the channel attention, spatial attention utilises global average pooling (AvgPool) and a SoftMax activation function to extract spatial information from the feature map [32, 33]. Average pooling performs equal-weighted integration of all activations, which more effectively maintains contextual features compared to max pooling's biased focus on local extremums [34]. The attention map is obtained by matrix multiplication with the compressed matrix. Following this, the input is subjected to the Sigmoid activation function, and then the attention map is multiplied by the input, thereby yielding a feature matrix with enhanced spatial features. The aforementioned process can be represented as follows:

$$R_3^{\text{spatial}} = R_3 \otimes \sigma \left( S_q(R_3) \times \text{SoftMax} \left( \text{AvgPool}(S_k(R_3)) \right) \right) \tag{6}$$

where $S_q$, and $S_k$ represent convolutional layers with kernel sizes of $1 \times 1 \times 1$, respectively. Finally, the channel attention features, and the spatial attention features are added together to obtain the final output of one RFEE layer. This can be represented as:

$$R_{output} = R_3^{\text{channel}} \oplus R_3^{\text{spatial}} \tag{7}$$

where $\oplus$ represents element-wise summation, and Channel and Spatial denote channel attention and spatial attention, respectively.

In the MGFA-Net configuration, there are five RFEE modules. The input of each module is denoted as $X_i$ ($i = 0,1,2,3,4$), with $X_0$ representing the input grayscale image. During the downsampling process, it is necessary to integrate the features from the myocardial region in the MRG module. The following equation represents this process:

$$X_i = X_i \oplus M_i \tag{8}$$

This operation is performed layer by layer, ultimately integrating features from each layer to enhance the model's expressiveness.

### 3.1.4 Multi-scale Feature Fusion Module

In the task of coronary artery segmentation, the deployment of MSFF is of great significance, primarily due to the significant size inconsistency and morphological irregularity of the coronary artery system [35]. The variation in size, shape, and distribution of coronary artery branches among different patients poses a significant challenge to traditional single-scale feature extraction methods, which are unable to accurately capture all the details. Furthermore, medical images frequently contain noise and artefacts [36], which serve to compound the complexity of the segmentation task. The MSFF module enhances the model's adaptability to coronary arteries of different scales and morphologies by progressively aggregating features from adjacent scales in the decoding path [37]. Rather than making direct multi-branch predictions on refined features, the MSFF module attains more efficient prediction results by gradually fusing features from adjacent scales in the decoding path. At each aggregation stage, the module dynamically integrates feature information from both the previous and current scales, adaptively extracting key features related to semantics while preserving spatial details. This progressive aggregation strategy has been shown to enhance the representation of multi-scale features and to effectively improve the model's ability to capture and predict complex structures, demonstrating stronger generalisation performance and stability [38, 39]. The MSFF is illustrated in Figure 5.

In MSFF, it is necessary to gradually aggregate features between adjacent scales. To extract features that are correlated between adjacent scales, the Adaptive Fusion Block (AFB) has been designed, as shown in Figure 5 (b). This block receives feature maps from different scales and uses feature concatenation (Concat), dimensionality reduction via $1 \times 1 \times 1$ convolution, and Sigmoid activation to generate attention weight matrices. These matrices are then used to weigh the input features. Subsequently, a $3 \times 3 \times 3$ convolution is applied to further extract multi-scale information and enhance the feature representation.

Assuming the inputs are four feature maps at different scales, denoted as $D_i$ ($i = 1,2,3,4$), the following steps are taken. First, $D_4$ passes through a $1 \times 1 \times 1$ convolutional operation to reduce the number of feature channels to 8 and then upsampled to match the size of $D_3$. Subsequently, the channel number of $D_3$ is also reduced to 8. The two feature maps with reduced channels are then fed into an AFB for feature fusion. This process is repeated after applying a upsampling layer. The final output of the model, $y_{output}$, can be represented as follows:

$$F_3 = f_u\big(f_c(D_4)\big) \tag{9}$$

$$F_2 = f_u\left(f_A\big(F_3, f_c(D_3)\big)\right) \tag{10}$$

$$F_1 = f_u\left(f_A\big(F_2, f_c(D_2)\big)\right) \tag{11}$$

$$y_{output} = f_c\left(f_A\big(F_1, f_c(D_1)\big)\right) \tag{12}$$

In this study, the upsampling operation is denoted by $f_u$, with the dimensions (C, H, W, D) of F3, F2, and F1 being (8, 40, 40, 32), (8, 80, 80, 64), and (8, 160, 160, 128), respectively. The convolutional operation with a kernel size of $1 \times 1 \times 1$ is represented by $f_c$, while $f_A(\cdot)$ indicates the AFB, which is employed to integrate features from disparate scales. The objective of the present study is to design an AFB that would effectively integrate different feature inputs and dynamically adjust the weight distribution of features through an attention mechanism. Assuming the inputs are $U$ and $F$, and the output is $a_3$, the process can be described as follows:

$$a_1 = Concat(U, F) \tag{13}$$
$$a_2 = \sigma\big(\text{Conv}_{1\times1\times1}(a_1)\big) \tag{14}$$
$$a_3 = \text{Conv}_{3\times3\times3}(a_1 \otimes F) \tag{15}$$

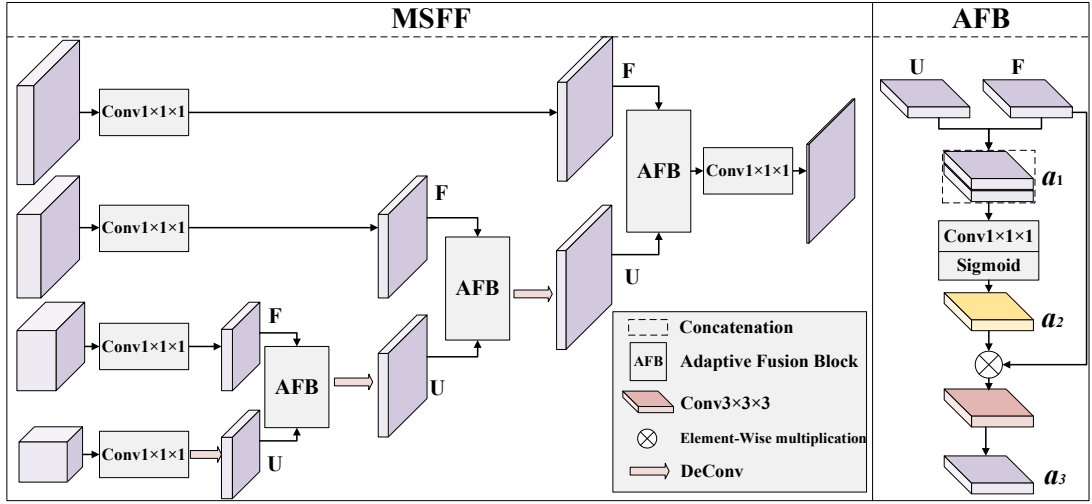

Figure 5. The architecture of the Multi-scale Feature Fusion Module

**3.1.5 Monte Carlo Dropout**

In order to measure the model segmentation uncertainty and to modestly improve the segmentation accuracy of the coronary arteries, this method introduces a Monte Carlo Dropout (MC Dropout), which estimates the model's prediction uncertainty by using multiple forward propagation in the inference phase [40]. During the training process, the model is combined with the standard Dropout technique to randomly discard neurons in the forward propagation, thus simulating different network structures, while in the inference phase, multiple predictions are obtained by applying the Dropout multiple times to perform Monte Carlo Sampling [41]. By calculating the mean and variance of these results, we are able to provide a measure of uncertainty for each prediction region, especially in difficult regions in coronary segmentation, where regions with higher uncertainty can provide additional diagnostic information to physicians.

In the specific context of coronary artery segmentation, during the inference stage, the objective is to obtain uncertainty estimates. To this end, Dropout is enabled and multiple forward passes are performed to simulate multiple network structures. Assuming G forward propagations are conducted, with G being set at 5, the prediction results will be obtained, represented by $\hat{y}_1, \hat{y}_2, \ldots, \hat{y}_G$. We will describe in the experimental section the values taken by G. Each $\hat{y}_k$ denotes the prediction map from the $k_{th}$ forward propagations. Each forward propagation generates a segmentation result, and the final prediction is obtained by calculating the mean of all forward pass results. The process can be described as follows:

$$\hat{y}_{\text{avg}} = \frac{1}{G} \sum_{k=1}^{G} \hat{y}_k \qquad (16)$$

where $\hat{y}_{\text{avg}}$ is the final average prediction map, which serves as the model's final output. The variance of each prediction is used as a measure of uncertainty, indicating the model's confidence in predicting a particular region, and can be described as:

$$\theta^2 = \frac{1}{G} \sum_{k=1}^{G} (\hat{y}_k - \hat{y}_{\text{avg}})^2 \qquad (17)$$

where $\theta^2$ denotes the variance, which quantifies the model's confidence in its predictions for each region. By calculating the variance, an uncertainty measure can be provided for each predicted pixel. In the task of coronary artery segmentation, the model may produce less accurate predictions for regions with high uncertainty (i.e., regions with higher variance), which are typically caused by ambiguities in the data or instabilities in the model. Clinicians can use this uncertainty information to pay additional attention to these regions.

**3.1.6 Loss function**

The Binary Cross-Entropy (BCE) loss function is used tool in the field of binary classification, serving to quantify the discrepancy between the anticipated outcomes and the actual labels. In medical image segmentation tasks, the BCE loss is frequently employed to quantify the probability discrepancy of each voxel belonging to the target class. The BCE loss formula is as follows:

$$\mathcal{L}_{bce} = -\frac{1}{N}\sum_{i=1}^{N}[y_i \log(\hat{y_i}) + (1-y_i)\log(1-\hat{y_i})] \tag{24}$$

The Dice loss function is a similarity measurement method that is frequently employed in the context of medical image segmentation. It quantifies the degree of overlap between the predicted segmentation region and the true segmentation region. The Dice coefficient, which ranges from 0 to 1, is a measure of how well the predicted and true regions overlap. Higher values indicate greater overlap between the predicted and true regions. The Dice loss function can be defined as the inverse of the Dice coefficient, as shown below:

$$\mathcal{L}_{dice} = 1 - \frac{2\sum_{i=1}^{N} y_i \hat{y_i} + \epsilon}{\sum_{i=1}^{N} y_i + \sum_{i=1}^{N} \hat{y_i} + \epsilon} \tag{25}$$

where $N$ is the total number of pixels, $y_i$ is the true label (0 or 1), $\hat{y_i}$ is the predicted probability value (0 to 1), and $\epsilon$ is a small constant to prevent division by zero.

In order to leverage the benefits of both loss functions, they can be amalgamated to form a composite loss function. This combination has the capacity to optimise the network's classification accuracy (through BCE loss) and the precision of the segmentation shape (through Dice loss) simultaneously. The composite loss function is typically represented as:

$$\mathcal{L}_{total} = \lambda \mathcal{L}_{dice} + (1-\lambda)\mathcal{L}_{bce} \tag{26}$$

where $\lambda$ is 0.5, used to balance the relative contributions of BCE loss and Dice loss. $\mathcal{L}_{bce}$ and $\mathcal{L}_{dice}$ are the Binary Cross Entropy loss and Dice loss, respectively.

**3.2 Coronary Artery Stenosis Detection**

**3.2.1 Centerline Extraction**

The extraction of arterial centerlines is crucial in CCTA imaging analysis, particularly in the study of arterial anatomy and the detection of arterial stenosis. The process of centerline extraction facilitates not only the accurate delineation of arterial morphology but also the acquisition of essential data for lesion assessment, surgical planning and interventional treatment. The primary objective is to eliminate redundant foreground pixels in the binary image while preserving the connectivity and topological structure of the vascular tree, thereby obtaining a simplified representation of the arterial segments.

During the centerline extraction process, it is essential to locate each point on the centerline at the geometric center of the arterial contour. Morphological algorithms are employed, which progressively dissect the structures within an image through a series of image processing operations, such as erosion and dilation. Specifically, the erosion operation is instrumental in reducing the vascular region by removing excess foreground pixels until the remaining image contains only the vessel's centerline. The detailed process is shown in Algorithm 1.

This process is ordinarily executed iteratively through erosion operations until the topological structure of the vascular tree remains unchanged and connectivity is preserved. This method accurately extracts vessel centerlines through multiple iterations, preserving important structural details. Morphological techniques effectively manage complex structures, like vessel branches and intersections, ensuring the centerline reflects the vessel's anatomical shape.

---

**Input:**
  X: Binary vessel image
  B: Structuring element (used for erosion, opening, etc.)
**Output:**
  Skeleton S (i.e., the vessel centerline)
1. Initialization:
  S = ∅
  Let temporary image T=X
2. Loop until T is empty or cannot be eroded further:
  3. Erode T to get T-eroded=Erode(T, B)
  4. Perform an opening operation on T-eroded to get T-opened=Open(T-eroded, B)
  5. Compute the skeleton fragment: temp=T−T-opened
  6. Merge the current skeleton fragment into S: S=S∪temp
  7. Update T=T-eroded
8. Exit the loop when T is empty or cannot be eroded further.
9. Return S (the final extracted vessel centerline).

---

Algorithm 1: Centerline extraction using the coronary artery segmentation results

**3.2.2 Individual artery separation**

The objective of this paper is to detect stenosis in arterial segments. During the detection process, centerline points were obtained in units of pixels, which were then connected to form a complete vascular tree structure. To elucidate the topological relationships of the arteries and decompose them into independent arterial segments, an edge-linking algorithm [42] was employed to identify and connect the edge points on the centerline. This process of segmentation resulted in the delineation of the vascular tree and is crucial for distinguishing different arterial segments.

Specifically, the centerline comprises three types of points according to graph theory. The first type consists of points with a degree of 1, which represent the endpoints of the arteries. The second type consists of points with a degree of 2, which represent connecting points. The third type encompasses points with a degree greater than 2, which represent branching points. Utilizing these categories of points allows the determination of the start and end points of coronary artery segments separation.

As shown in Figure 6, the branches of the arteries and the branching points on their centerlines are described.

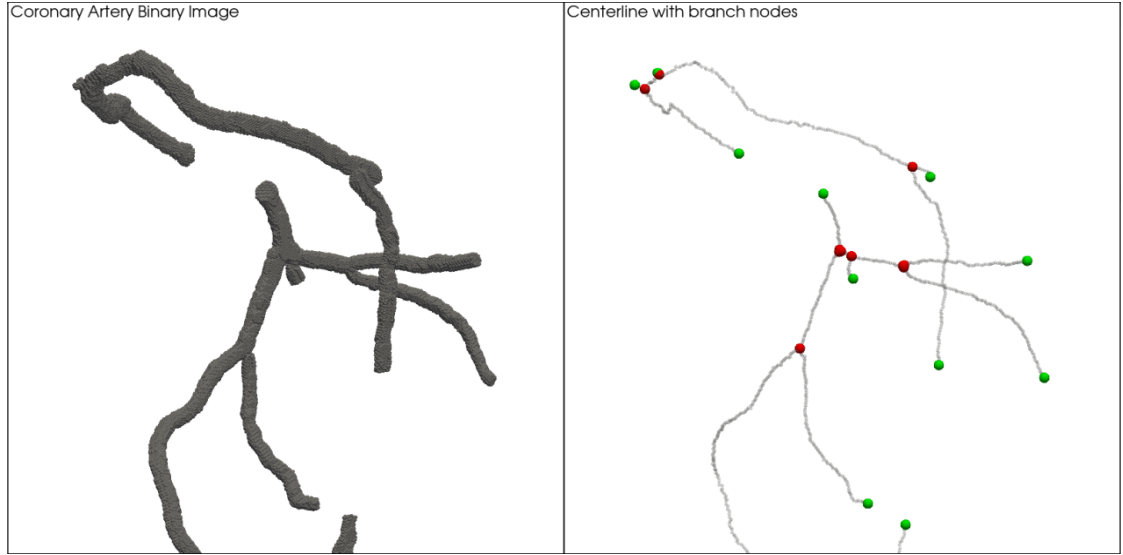

Figure 6 Branching points on the centerline of the artery. The left image shows the binary map of the coronary arteries, while the right image displays the centerlines of the coronary arteries and their branching nodes. In the right image, green points represent nodes with a degree of 1, red points represent nodes with a degree of 3 or higher, and gray points represent nodes with a degree of 2.

**3.2.3 Calculation of Arterial Cross-sectional Area**

To calculate the cross-sectional area of arterial segments using the centerline proximity region, it is necessary to sort the centerline points of the coronary artery segments from the proximal to the distal end. Assuming the necessity to calculate the cross-sectional area at the p-th point($p$) on the centerline, the direction vector v is a 3D vector that represents the local tangent direction of the artery's centerline at the $p$. The direction vector $v = (v_x, v_y, v_z)$ between points $p_{-10}$ and $p_{+10}$ is first computed, where 10 represents the number of centerline points, (x,y,z) represent the three directions in 3D space, $v_x = p_{+10}.x - p_{-10}.x, v_y = p_{+10}.y - p_{-10}.y, v_z = p_{+10}.z - p_{-10}.z.$. Assuming point $m$ is within region s and belongs to the label, set $\mathbf{thrd}$ as the dot product threshold. If $|(p - m) \cdot v| < \mathbf{thrd}$ at this point, then $m$ is identified as a pixel belonging to the cross-section. Subsequently, the count of qualifying points within radius $s$ can be computed. The specific process can be represented as Algorithm 2, which is used to calculate the number of pixels of the arterial cross-sectional area for each centerline point.

Following the acquisition of the pixels on the cross-section, it is necessary to calculate their actual physical size. The pixel spacing on the cross-section is contingent on both the pixel spacing and the direction of the cross-section's normal. Initially, we compute the two basis vectors $b_1$ and $b_2$ of the cross-section as the direction vectors of the cross-section plane. Subsequently, utilising the 3D voxel spacing, these two basis vectors are employed to calculate the 2D pixel spacing on the cross-section, denoted as $p_1$ and $p_2$. Finally, the number of cross-sectional pixels is multiplied by $p_1$ and $p_2$ to obtain the cross-sectional area A. The detailed process is delineated in Algorithm 3, which provides a method for calculating the cross-sectional area of arteries.

**Input:**
    *p*: center point
    *centerline_points*: set of centerline points
    *s*: search radius
    *thrd*: dot product threshold
    *label*: binary vessel image

**Output:**
    *Points*: Number of cross-sectional points

1. Get the points 10 positions before and after point *p*:
$$p_{-10} = \text{centerline\_points}[p - 10]$$
$$p_{+10} = \text{centerline\_points}[p + 10]$$
2. Calculate the directional vector $\boldsymbol{v}$ between $p_{-10}$ and $p_{+10}$:
3. Initialize the cross-sectional point set $C = [\ ]$
4. For point $m \in$ label within radius $s$:
   a. Calculate the L:
$$L = \boldsymbol{v} \cdot (m - \boldsymbol{p})$$
   b. If $|L| < \boldsymbol{thrd}$, then consider the point to be on the cross-section:
    Add $m$ to the cross-sectional point set $C$
5. Calculate the number - of cross-sectional points:
   **Points** $= |C|$

Algorithm 2: Calculate the number of pixels of the arterial cross-sectional area for each centerline point

**Input:**
*v*: Cross-section normal vector
*points*: Number of cross-sectional pixels
*dx*, *dy*, *dz*: Voxel spacing in x, y, and z directions

**Output:**
*A*: Cross-sectional area

1. Generate two basis vectors $b_1$ and $b_2$:
  1.1 Use an initial vector [1,0,0], and compute the cross product with $\boldsymbol{v}$ to obtain $b_1$.
$$b_1 = v \times [1,0,0]$$
  1.2 Compute the cross product of $v$ and $b_1$ to obtain $b_2$.
$$b_2 = v \times b_1$$
2. Calculate the actual lengths of the basis vectors in 3D space:
   Multiply the components of $b_1$ and $b_2$ in the $x$, $y$, and $z$ directions by $\boldsymbol{dx}$, $\boldsymbol{dy}$ and $\boldsymbol{dz}$, respectively, to obtain their physical components $p_1$ and $p_2$.
$$p_1 = \sqrt{(b_{1x} \cdot d_x)^2 + (b_{1y} \cdot d_y)^2 + (b_{1z} \cdot d_z)^2}$$
$$p_2 = \sqrt{(b_{2x} \cdot d_x)^2 + (b_{2y} \cdot d_y)^2 + (b_{2z} \cdot d_z)^2}$$
3. Calculate the cross-sectional area:
   Multiply the number of cross-sectional pixels points by the magnitudes of $p_1$ and $p_2$ to obtain the actual cross-sectional area A.
$$A = \boldsymbol{Points} \times p_1 \times p_2$$

Algorithm 3: Calculate the actual cross-sectional area of arteries

### 3.2.4 Arterial Segment Stenosis Detection

In the stenosis detection task, each coronary artery segment is analysed independently. Following the acquisition of the cross-sectional area at each position of the coronary artery segment, points are defined as local minima if their second derivative is -2, and as local maxima if their second derivative is 2.

The candidate point list is formed by all local minima, and the minimum cross-sectional area of the artery segment is defined as the smallest among the candidate points, denoted as $A_{\min}$. In a similar manner, all local maxima form another candidate point list, and the largest cross-sectional area among these candidate points is defined as the maximum cross-sectional area of the artery segment, denoted as $A_{\text{ref}}$.

**Input:**
$I$: Binary segmentation result of the coronary arteries
$d$: Voxel spacing
**Output:**
$p$: Degree of stenosis and coordinates of stenosis location
1. Skeletonize $I$ to obtain the centerline.
2. Extract bifurcation and endpoint points from the centerline to segment the coronary arteries into artery segments.
3. Calculate the cross-sectional area of the coronary artery at the corresponding pixel positions along the centerline of each segment.
4. Remove artery segments shorter than 20 pixels.
For each artery segment:
1. Compute the second derivative sequence of the artery segment to find local minima and maxima points.
2. Calculate the degree of stenosis using the formula.
3. If the degree of stenosis is greater than 0.1, add the point to $p$.

Algorithm 4: Arterial stenosis detection

According to the coronary artery stenosis grading criteria recommended by the Society of Cardiovascular Computed Tomography, coronary artery stenosis is divided into four levels: mild (1%-24%), moderate (25%-49%), severe (50%-69%), and very severe (70%-100%). In this study, we adopted the same grading criteria for coronary artery stenosis. The percentage of stenosis in an artery segment can be expressed as shown in Eq. 18.

$$b = \left(1 - \frac{A_{\min}}{A_{\text{ref}}}\right) \times 100\% \qquad (18)$$

If the length of a short artery is less than 20 pixels, it is removed. If there are no points with a second derivative equal to -2, it is assumed that there are no stenosis points within the artery segment. If there are no points with a second derivative equal to 2, the maximum diameter in the diameter sequence along the centerline is selected as $A_{\text{ref}}$. Algorithm 4 describes the entire stenosis detection algorithm. The visualization of the entire stenosis detection process is shown in Figure 7.

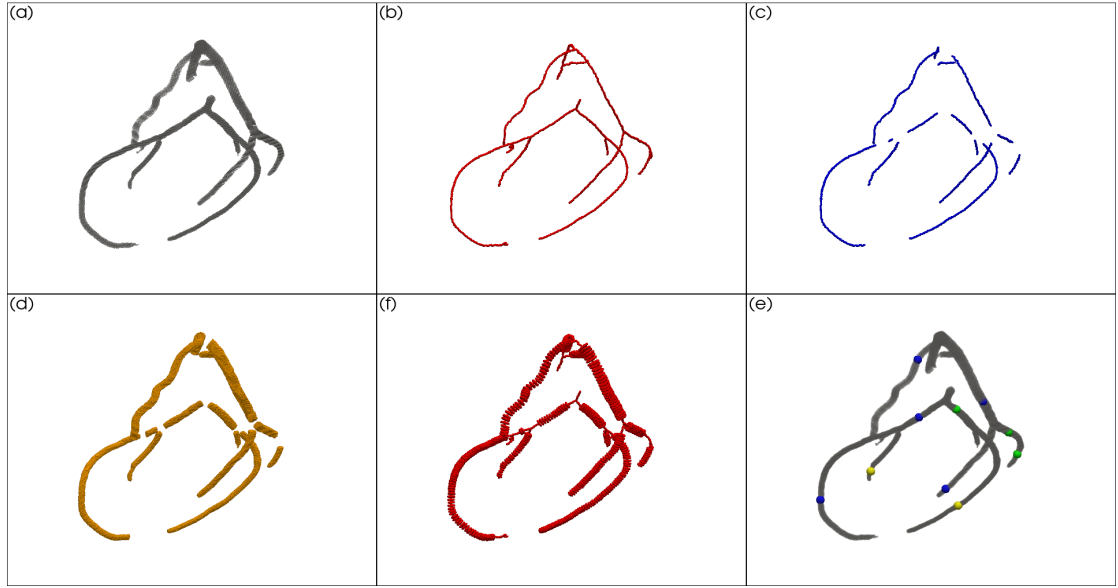

Figure 7: Stenosis detection process. **(a)** represents the segmentation result, **(b)** represents the centerline skeleton derived from the segmentation result, **(c)** represents the segmented centerline segments, **(d)** represents the segmented arterial segments, **(e)** represents the visualization of cross-sectional areas measured at various locations (with an interval of 3), and **(f)** represents the final stenosis degree and stenosis location obtained by calculating the cross-sectional areas (green for minimal, blue for mild, yellow for moderate, and red for severe).

### 3.3 Evaluation metrics

### 3.3.1 Evaluation of Artery Segmentation

In order to evaluate the performance of the segmentation, four metrics were utilised. Initially, three classic quantitative evaluation metrics were employed: The Dice Similarity Coefficient (DSC), recall, and precision. The range of values for these three metrics is [0, 1], with higher values indicating better segmentation quality. In addition, given the critical importance of vascular boundaries in coronary artery segmentation results, the Hausdorff Distance (HD) was introduced as an evaluation metric in the quantification process. The HD is employed to measure the maximum distance between the boundaries of the reference object and the automatically segmented object, and it is defined as follows:

$$\mathrm{HD}(X, Y) = \max \left\{ \sup_{x \in X} \inf_{y \in Y} d(x, y), \sup_{y \in Y} \inf_{x \in X} d(x, y) \right\} \quad (19)$$

$X$ and $Y$ represent the sets of reference boundary points and automatically segmented boundary points, respectively. $d(x, y)$ is the Euclidean distance, defined as: $d(x, y) = \sqrt{(x_a - y_a)^2 + (x_b - y_b)^2 + (x_c - y_c)^2}$, $sup$ denotes the supremum (least upper bound), and $inf$ denotes the infimum (greatest lower bound). The Hausdorff Distance is a metric that is employed to evaluate the discrepancy between the boundaries of the segmentation results and the reference standard. This is achieved by calculating the maximum of the minimum distances between two sets of boundary points. A smaller value indicates that the segmentation result is closer to the true boundary. Among these metrics, the DSC primarily measures the overlap of pixels between the estimated result and the ground truth and is thus the most important evaluation metric in segmentation tasks. In this study, we prioritise the performance ranking based on the DSC results, followed by the 95% Hausdorff Distance (95% HD), recall, and precision.

**3.3.2 Stenosis Detection Evaluation**

It is imperative to establish a set of rules to achieve objective and automated matching, given the potential incongruity between the stenosis points detected from the ground truth and the predicted arterial contours.

**Rule 1:** Initially, the predicted arterial contour must match with the arterial segment in the ground truth. Subsequently, the distances between the two endpoints of each arterial segment must be calculated. If both distances are found to be less than the ***detection radius*** (a hyperparameter), the segments are deemed to have been successfully matched. In the event of two segments being matched, the stenosis points within the matched segments are marked as matched points.

**Rule 2:** In the absence of a matching arterial segment in the ground truth, the second rule is to be applied. Within a circle of radius equal to the ***detection radius*** around the predicted arterial segment, the nearest stenosis point in the ground truth should be selected as the corresponding matched stenosis point for the predicted arterial segment.

In order to quantitatively evaluate the performance of the stenosis detection algorithm, it is necessary to clarify several key definitions.

1. In the event of the detection of matching stenosis points in both the predicted arterial contour and the ground truth, these are defined as True Positive (TP) samples.

2. Conversely, if a stenosis point is detected in the ground truth but not in the predicted arterial contour, it is defined as a False Negative (FN) sample.

3. A detected stenosis in the predicted arterial contour is a False Positive (FP) if no corresponding stenosis exists in the ground truth annotation.

Similar to binary classification tasks, we use the True Positive Rate (TPR) and Positive Predictive Value (PPV) to evaluate the performance of the stenosis detection algorithm [43]. Their definitions are as follows:

The True Positive Rate (TPR), also known as recall, is defined as:

$$\text{TPR} = \frac{\text{TP}}{\text{TP} + \text{FN}} \tag{20}$$

The Positive Predictive Value (PPV), also known as precision, is defined as:

$$\text{PPV} = \frac{\text{TP}}{\text{TP} + \text{FP}} \tag{21}$$

Furthermore, the Absolute Root Mean Squared Error (ARMSE) and the Relative Root Mean Squared Error (RRMSE) are utilised to evaluate the degree of stenosis detected in TP samples. The definitions of ARMSE and RRMSE are as follows:

Absolute Root Mean Squared Error (ARMSE):

$$\text{ARMSE} = \sqrt{\frac{1}{N} \sum_{i=1}^{N} \left(A_i^{\text{pred}} - A_i^{\text{true}}\right)^2} \tag{22}$$

Relative Root Mean Squared Error (RRMSE):

$$\text{RRMSE} = \sqrt{\frac{1}{N}\sum_{i=1}^{N}\left(\frac{A_i^{\text{pred}} - A_i^{\text{true}}}{A_i^{\text{true}}}\right)^2} \tag{23}$$

where $N$ represents the number of True Positive (TP) samples, $A_i^{\text{pred}}$ is the predicted cross-sectional area of the $i$-th sample, and $A_i^{\text{true}}$ is the true cross-sectional area of the $i$-th sample.

## 4 Experiment

### 4.1 DataSet

A public large-scale dataset, ImageCAS [44], consisting of 1,000 CCTA images acquired using a Siemens 128-slice dual-source CT scanner, is utilized for coronary artery segmentation. Each 3D CCTA image has a size of 512×512×(206-275) pixels, with an in-plane resolution of 0.29-0.43 mm² and a slice spacing of 0.25-0.45 mm. The left and right coronary arteries were independently annotated by two radiologists and underwent cross-validation. The images in the dataset were obtained from various CT scanners, including the Revolution CT by GE Healthcare and the SOMATOM Definition Flash by Siemens Healthcare, with scans performed by certified radiologic technologists. The pixel spacing ranges from 0.28 to 0.41 mm, with slice thicknesses of 0.5 to 1.0 mm, and the number of slices per scan varies between 210 and 275. The age of the patients ranges from 46 to 78 years.

### 4.2 Experimental Setup

The network was implemented using the PyTorch framework and trained on an NVIDIA RTX 4090 GPU. The Adam optimiser was utilised with an initial learning rate of 0.002, and a cosine annealing learning rate scheduler was employed. The model was trained for a total of 200 epochs. To enhance the model's generalisation capability and prevent overfitting, all data were normalised and augmented prior to training. Data augmentation techniques employed included horizontal flipping and random cropping, with the cropped image size set to 128×160×160. The 1,000 cases were randomly divided into three sets: 700 were allocated for training, 100 for validation, and 200 for testing.

### 4.3 Artery Segmentation Results

#### 4.3.1 Comparison experiments

A comparison of four methods was conducted, including 3D U-Net [10], V-Net [11], DenseVoxNet [12], and CS²-Net [13]. To ensure fairness in the comparison, it was imperative that all experiments followed the same preprocessing pipeline, and all models were retrained using a consistent learning strategy to achieve optimal performance.

As demonstrated in Table 1, MGFA outperformed the four compared models. The model achieved 85.04% in DSC, 84.24% in Precision, 86.27% in Recall, and 6.1294 mm in HD95. Specifically, MGFA surpassed U-Net 3D, DenseVoxelNet, V-Net, and CS2-Net by 3.57%, 2.41%, 2.1%, and 1.75% in DSC, respectively. This finding suggests that MGFA demonstrated superiority in capturing vascular details during the segmentation task, particularly in accurately delineating stenotic regions and efficiently reducing over-segmentation in regions external to vessels. To facilitate intuitive comparison, the experimental results were visualised. The visualisation results are displayed in Figure 8, which presents the segmentation results of two CTA images selected from the dataset. The figure reveals that some methods suffered from over-segmentation and inaccurate segmentation of stenotic regions. In contrast, MGFA demonstrated a capacity to overcome these issues to a certain extent.

Tabel 1. Comparison of Coronary Artery Segmentation Performance on ImageCAS Dataset using different models

| Network | Dice(%) | Precision(%) | Recall(%) | HD95(mm) |
| --- | --- | --- | --- | --- |
| Unet 3D[10] | 81.47 | 79.69 | 83.95 | 15.9125 |
| DenseVoxelNet[12] | 82.63 | 83.53 | 82.19 | 12.0250 |
| V-net[11] | 82.94 | 81.23 | 85.28 | 12.8953 |
| $CS^2$-Net[13] | 83.29 | 82.41 | 84.68 | 8.1108 |
| **MGFA(Ours)** | **85.04** | **84.24** | **86.27** | **6.1294** |

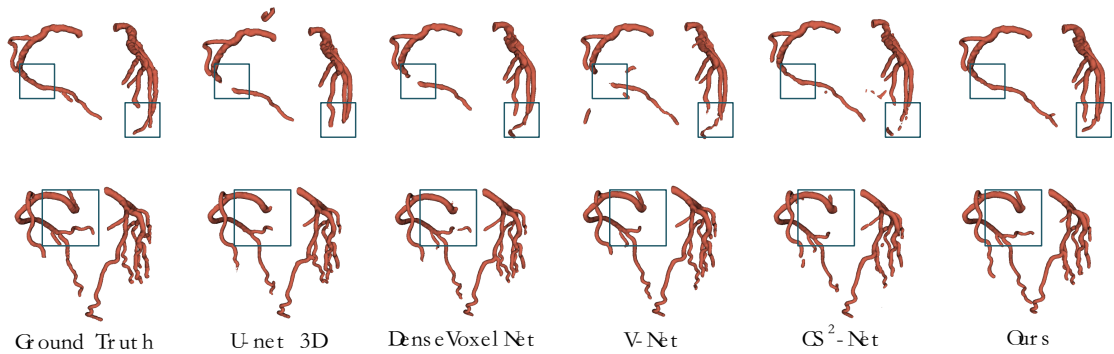

Ground Truth　　U-net 3D　　DenseVoxel Net　　V-Net　　$CS^2$-Net　　Ours

Figure 8 Visual comparison of two cases of segmentation using different models

### 4.3.2 Ablation studies

To validate the effectiveness of each module in MGFA, a comprehensive analysis was conducted on the following modules: MRG, MSFF and RFEE. Utilising 3D U-Net as the baseline, the performance of each module was evaluated based on metrics such as Dice, HD95, Recall, and Precision.

As illustrated in Table 2, the quantitative segmentation outcomes for diverse modules are presented, and the enhancement achieved by each module is evident. In the table, model 1 incorporates the MRG module on the basis of the Baseline model. The addition of the MRG module increases the DSC index of the model by 1.80%, and the HD95 reaches 8.2037, indicating a more substantial improvement. The MRG employs the myocardial expansion region as a guide to minimise the segmentation area to within the myocardial expansion region, thereby reducing over-segmentation outside this region. The segmentation outcomes are presented in Figure 9, model 1, demonstrating a reduction in superfluous segmentation relative to the baseline. Furthermore, the incorporation of the RFEE module and MSFF module on top of the baseline in Models 2 and 4 resulted in an improvement in the DSC index by 1.03% and 1.01%, respectively. Similarly, the incorporation of the RFEE module on top of Model 1 in Model 3 led to an enhancement in the DSC index by 0.44%. Finally, the integration of the RFEE module on top of Model 1 in Model 5 demonstrated a 0.44% improvement in the DSC index. The RFEE model fuses the spatial and channel features of the image, thereby enhancing the model's segmentation performance. The segmentation outcomes are presented in Figure 9, model 3, which demonstrates an enhancement in the segmentation accuracy of the arterial fine branches in comparison to model 1. Model 5 incorporates the MSFF module on the foundation of model 1, resulting in an improvement of 0.26% in the DSC compared to model 1. The MSFF module fuses multiscale features, thereby refining the segmentation results and enhancing segmentation accuracy. The segmentation results, as illustrated in Figure 9, demonstrate that model5 improves the segmentation accuracy at the terminal points of arteries relative to model1. Furthermore, the combination of the MRG, MSFF and RFEE modules with Baseline has been shown to enhance the DSC index of segmentation, attributable to the synergistic effect of the three modules. This results in an improvement of 3% in the DSC compared with Baseline. Furthermore, the MC Dropout method is employed to quantify the uncertainty of the MGFA segmentation results, thereby enhancing the segmentation effect. A comparative analysis reveals that the MGFA incorporating uncertainty quantization enhances the DSC metric by 0.57% when compared with the MGFA devoid of uncertainty quantization.

Tabel 2 Comparison of Coronary Artery Segmentation Performance of ImageCAS Dataset in Different Ablation Study Models

| Network | MRG | MSFF | RFEE | Dice(%) | Precision(%) | Recall(%) | HD95(mm) |
|---|---|---|---|---|---|---|---|
| Baseline | | | | 81.47 | 79.69 | 83.95 | 15.9125 |
| Model 1 | √ | | | 83.27 | 82.66 | 84.42 | 8.2037 |
| Model 2 | | | √ | 82.50 | 81.55 | 83.96 | 11.1271 |
| Model 3 | √ | | √ | 83.71 | 82.07 | 85.88 | 7.3489 |
| Model 4 | | √ | | 82.48 | 81.34 | 84.21 | 12.5078 |
| Model 5 | √ | √ | | 83.53 | 81.84 | 85.78 | 8.6876 |
| Model 6 | √ | √ | √ | 84.47 | 83.75 | 85.57 | 6.3810 |
| Model 7 | **MGFA(UQ)** | | | **85.04** | **84.24** | **86.27** | **6.1294** |

As demonstrated in Figure 9, the visualisation outcomes of the ablation experiments are presented. It is evident that the incorporation of MRG, MSFF, and RFEE enhances the segmentation performance of the model. Specifically, the addition of MRG effectively reduces over-segmentation, while the integration of MSFF and RFEE renders the segmentation network more sensitive to small vessel regions, resulting in more accurate segmentation.

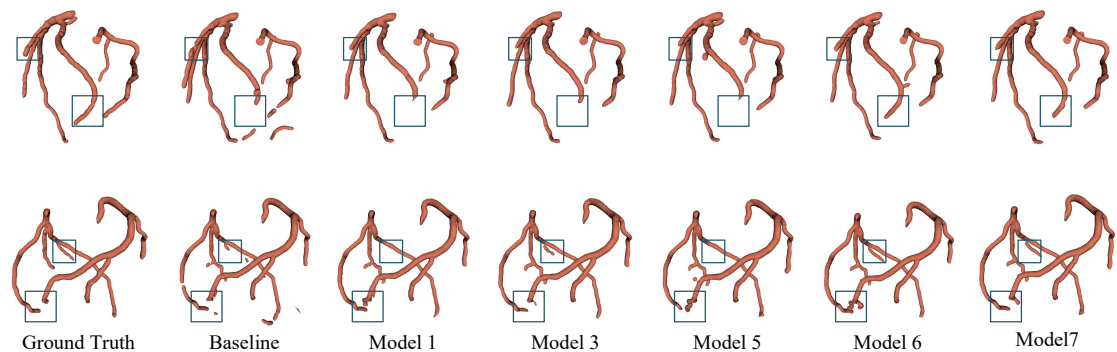

Ground Truth    Baseline    Model 1    Model 3    Model 5    Model 6    Model7

Figure 9 Comparison of two case visualizations of different segmentation models in ablation studies

As illustrated in Figure 10, the uncertainty quantization map, segmentation results and

Ground Truth are presented. The uncertainty quantization map is derived from the MC Dropout probability variance map through a process of normalization, with larger values indicating higher uncertainty. The uncertainty quantization map thus provides a reference and suggestion for the doctor to segment the uncertainty region in the actual diagnosis, thus facilitating the correct decision.

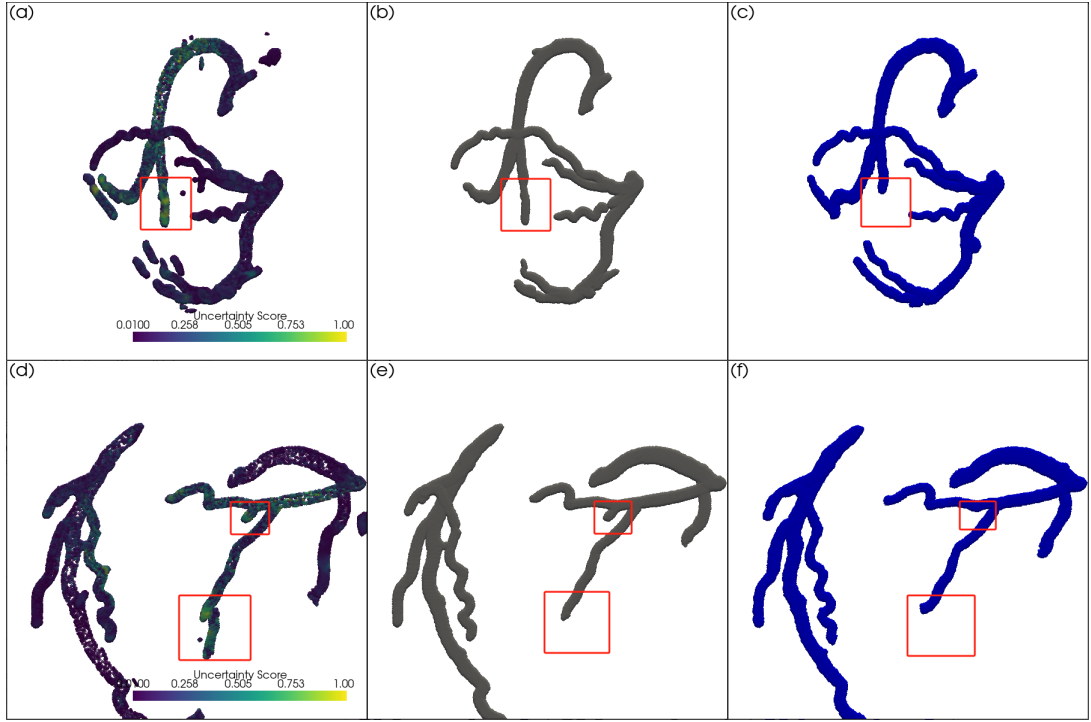

Figure 10: Uncertainty quantification map, (a) and (d) show the visualization of the uncertainty quantification score, where brighter colors indicate higher uncertainty in the region. (b) and (e) represent the segmentation result. (b) and (e) represent the Ground truth.

In Table 3, the number of inference times M used for multiple uncertainty quantification was tested, and it was determined that 5 was the optimal number. During the process of model inference, substantial overlaps were established across the entire image, resulting in regions near the image's center undergoing multiple inference times.

Tabel 3 The ultimate segmentation outcome for the quantity of uncertain inference times, designated as G.

| G | Dice(%) | Precision(%) | Recall(%) | HD95(mm) |
| --- | --- | --- | --- | --- |
| 10 | 84.99 | 83.43 | **86.54** | 6.6122 |
| 15 | 85.01 | 83.99 | 86.53 | 6.5591 |
| 20 | 84.96 | 83.99 | 86.46 | 6.8059 |
| 5 | **85.04** | **84.24** | 86.27 | **6.1294** |

To validate the performance of spatial and channel attention parallel strategies in RFEE, a comparison was made between the serial and parallel methods of spatial and channel attention. As demonstrated in Table 4, the parallel strategy of spatial and channel attention yielded the most optimal outcomes in RFEE.

Tabel 4 Model 1 represents spatial attention followed by channel attention, Model 2 represents channel attention followed by spatial attention, and Model 3 represents parallel attention

| Network | Dice(%) | Precision(%) | Recall(%) | HD95(mm) |
|---|---|---|---|---|
| Model 1 | 84.37 | 83.43 | 85.81 | 6.9380 |
| Model 2 | 84.13 | 81.41 | **87.52** | 7.3501 |
| **Model 3(Ours)** | **84.47** | **83.75** | 85.57 | **6.3810** |

In order to balance the loss function and obtain better segmentation, we conducted several experiments to select the value of λ. The results are shown in Table 5.

Table 5 Effect of parameter λ in the loss function

| λ | Dice(%) | Precision(%) | Recall(%) | HD95(mm) |
|---|---|---|---|---|
| 0.1 | 84.16 | 83.33 | 85.52 | 8.4354 |
| 0.2 | 84.27 | 83.27 | **85.76** | 7.9318 |
| 0.3 | 84.19 | 83.27 | 85.64 | 7.2323 |
| 0.4 | 83.62 | **84.23** | 83.58 | 13.4594 |
| 0.5 | **84.47** | 83.75 | 85.57 | **6.3810** |
| 0.6 | 84.38 | 83.78 | 85.51 | 6.3934 |
| 0.7 | 83.14 | 81.27 | 85.64 | 13.0312 |
| 0.8 | 83.77 | 83.03 | 85.02 | 9.5973 |
| 0.9 | 83.78 | 82.89 | 85.24 | 11.7111 |

**4.4 Artery Stenosis Detection Results**

In the experimental phase, the hyperparameter for the ***detection radius*** was set to 20. The subsequent figures illustrate a number of instances of stenosis detection. Figures 11 and 12 illustrate the Ground Truth of the coronary arteries, the stenosis detection results from the Ground Truth, the model's segmentation results, and its stenosis detection results, with the vascular segmentation model being MGFA

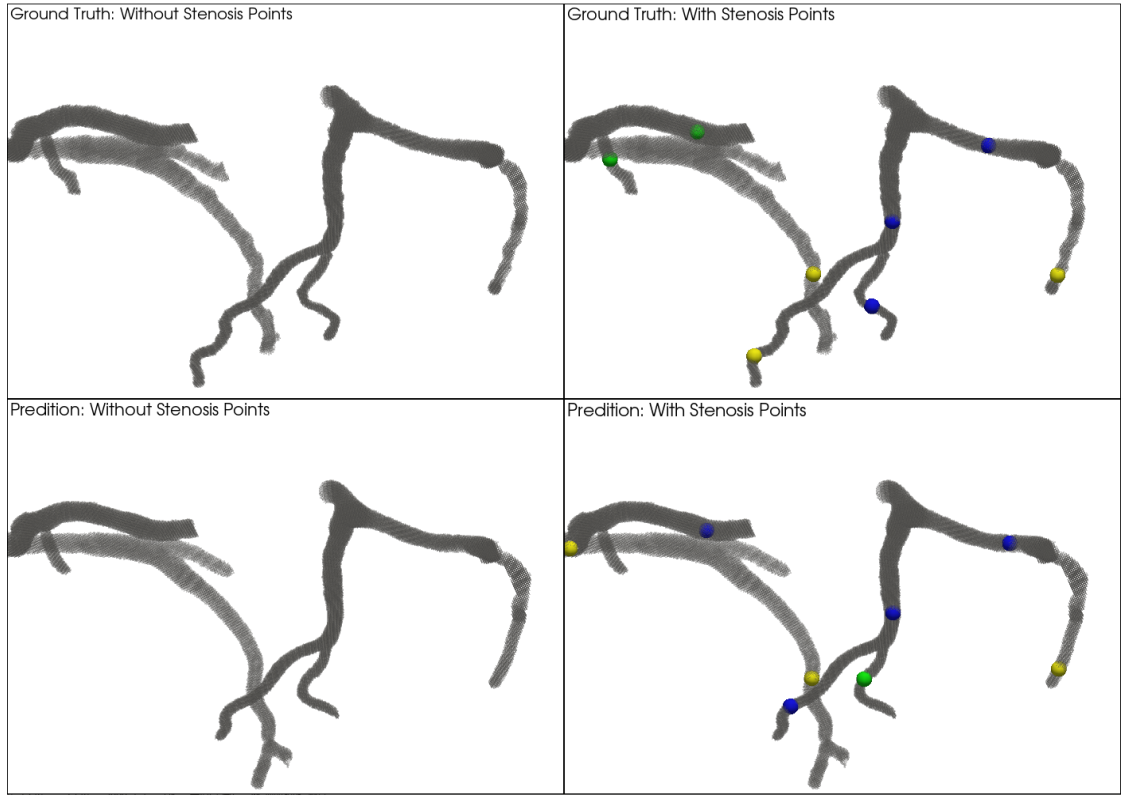

Figures 11 A case comparison of stenosis detection on ground truth as well as segmentation prediction (green for minimal, blue for mild, yellow for moderate, red for severe).

In order to provide quantitative evidence of the performance of the stenosis detection algorithm, the TPR, PPV, ARMSE and RRMSE were evaluated, as defined in the relevant formulas. The measurements from the Ground Truth were used as the standard in the evaluation. Utilising the coronary artery segmentation maps generated by MGFA, the stenosis detection algorithm detected 958 true positive (TP) points, 1141 false positive (FP) points, and 1039 false negative (FN) points from 200 images. The ensuing table illustrates the stenosis detection outcomes for the arterial contours derived from diverse models. According to Table 5, the stenosis detection algorithm using the segmentation results from MGFA has an ARMSE of 0.1537, an RRMSE of 0.3865, a TPR of 0.4564, and a PPV of 0.4797. A comparison of MGFA with U-net 3D reveals that MGFA enhances TPR and PPV by 2.34% and 5.46%, respectively, while concurrently reducing ARMSE and RRMSE by 0.0037 and 0.0201, respectively. In addition, MGFA demonstrated superiority over DenseVoxelNet in terms of TPR, PPV, ARMSE and RRMSE, with respective enhancements of 2.96% and 3.53%. Furthermore, MGFA exhibited a 0.008 decrease in ARMSE and a 0.0117 decrease in RRMSE when compared to V-net. A comparison of MGFA and $CS^2$-Net revealed enhancements of 1.9% and 2.2% in TPR, with corresponding reductions of 0.0087 and 0.0207 in PPV, ARMSE and RRMSE, respectively. The stenosis detection results are shown in Table 6.

Table 6 Comparison of stenosis detection performance of different segmentation model results

| Segmentation Model | Stenosis Type | TPR | PPV | ARMSE | RRMSE |
|---|---|---|---|---|---|
| U-net 3D | All | 0.4330 | 0.4251 | 0.1574 | 0.4064 |
|  | minimal | 0.4774 | 0.4842 | 0.1626 | 0.3409 |
|  | mild | 0.4240 | 0.4328 | 0.1319 | 0.3447 |
|  | moderate | 0.4298 | 0.3891 | 0.1710 | 0.5054 |
|  | severe | 0.3243 | 0.3918 | 0.3504 | 0.9027 |
| DenseVoxelNet | All | 0.4268 | 0.4444 | 0.1617 | 0.4042 |
|  | minimal | 0.4427 | 0.4220 | 0.1581 | 0.3162 |
|  | mild | 0.4267 | 0.4097 | 0.1338 | 0.3457 |
|  | moderate | 0.4171 | 0.4700 | 0.1921 | 0.5418 |
|  | severe | 0.3919 | 0.4594 | 0.3199 | 0.7046 |
| V-net | All | 0.4374 | 0.4630 | 0.1605 | 0.3989 |
|  | minimal | 0.4462 | 0.4691 | 0.1549 | 0.3160 |
|  | mild | 0.4327 | 0.4416 | 0.1385 | 0.3488 |
|  | moderate | 0.4531 | 0.4027 | 0.1552 | 0.3811 |
|  | severe | 0.3649 | 0.3919 | 0.4122 | 1.1696 |
| $CS^2$-Net | All | 0.4351 | 0.4577 | 0.1557 | 0.4072 |
|  | minimal | 0.4527 | 0.4642 | 0.1624 | 0.3349 |
|  | mild | 0.4349 | 0.4135 | 0.1232 | 0.3168 |
|  | moderate | 0.4187 | 0.4279 | 0.1907 | 0.5720 |
|  | severe | 0.4324 | 0.4595 | 0.2834 | 0.7553 |
| **MGFA-Net** | All | **0.4564** | **0.4797** | **0.1537** | **0.3865** |
|  | minimal | 0.4414 | 0.4684 | 0.1614 | 0.3077 |
|  | mild | 0.4618 | 0.4407 | 0.1215 | 0.3067 |
|  | moderate | 0.4638 | 0.4819 | 0.1849 | 0.5077 |
|  | severe | 0.4189 | 0.4729 | 0.2983 | 0.8471 |

**5 Discussion**

**5.1 Performance Analysis of Artery Segmentation**

In Figures 8 and 9, a visual comparison is presented of the results obtained by the various models in the ablation experiments and the comparison tests. The red-boxed area in the figures can be used to compare with other experimental models, and it is evident that the proposed MGFA has advantages in terms of stenosis segmentation and reduction of over-segmentation. The 95% Hausdorff Distance was utilised as a metric to evaluate the quality of segmentation in Tables 1 and 2, and MGFA-Net demonstrated an exceptional performance with a value of 6.1294 mm. This not only validates the segmentation results but also highlights the significance of the 95% Hausdorff Distance in subsequent stenosis detection. This finding underscores the efficacy of the myocardial region-guided MGFA-Net in confining the segmentation results within a specified range, thereby averting superfluous segmentation.

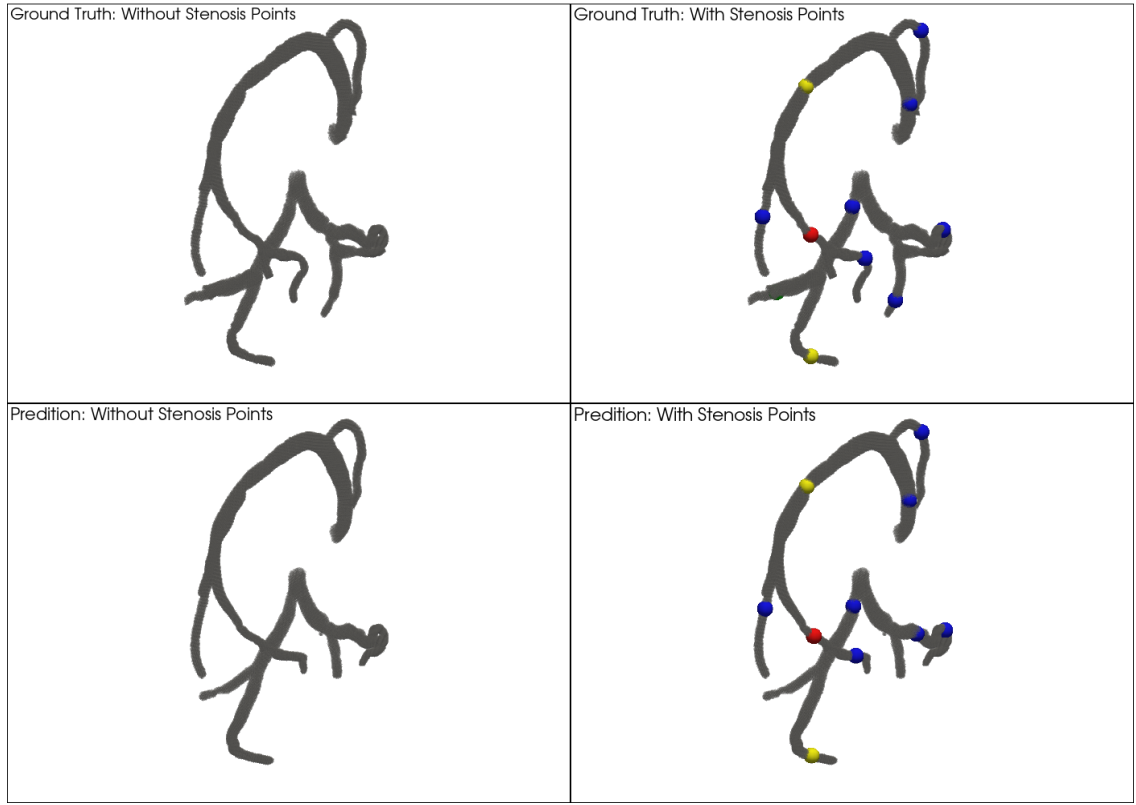

Figures 12 A case comparison of stenosis detection on ground truth as well as segmentation prediction (green for minimal, blue for mild, yellow for moderate, red for severe).

**5.2 Performance Analysis of Stenosis Detection**

As illustrated in Table 6, the proposed MGFA-Net demonstrates superior performance in terms of TPR, PPV, ARMSE, and RRMSE when compared to models such as U-Net. The stenosis detection algorithms achieved good performance in terms of positive predictive value (PPV) for moderate and severe stenosis by using the MGFA profile. Conversely, the PPV values for detecting moderate and severe stenosis using U-Net and V-Net were comparatively low, in contrast to other types of stenosis. In clinical practice, moderate and severe stenoses are of greater clinical significance and usually require intervention. Consequently, the arterial profile generated by MGFA is accurate and beneficial for stenosis detection. Illustrative examples of stenosis detection are presented in Figure 11 and Figure 12. The proposed method demonstrates enhanced accuracy in detecting moderate and severe stenosis (yellow and red). However, the performance of stenosis detection is less effective for mild stenosis and sites where incomplete arterial segmentation is present.

**5.3 Limitations**

Coronary artery segmentation continues to be an area of research with significant limitations. CTA is three-dimensional data, and the current models are highly complex, requiring a significant amount of computational resources during training and inference. The intricate geometry and structural complexity of coronary arteries pose a considerable challenge to even the most seasoned physicians in accurately delineating them. The presence of numerous small branches within coronary arteries continues to present a significant challenge to contemporary deep learning-based segmentation methods. As demonstrated in Figures 8 and 9, the model exhibits shortcomings in delineating these minute branches, and the overall segmentation outcome is not deemed optimal. It is important to note that the incorporation of each module into the model will result in an increase in model parameters and computational complexity. This, in turn, will impose further constraints on the segmentation task due to limited computational resources. In light of these challenges, we propose that in the future, we may consider incorporating the coronary topology into the segmentation network or utilising information such as the topology of the centerline to enhance the precision of the coronary artery endings with less computational complexity.

The location of stenosis detection is contingent on the completeness of the arterial contour segmentation, and is highly sensitive to the shape and smoothness of the segmentation results. Consequently, minor variations can result in erroneous detection locations. To illustrate this point, consider the presence of a minor branch within an arterial segment in the ground truth. In such a scenario, the arterial segment is typically divided into two for separate processing. In the event that this small branch is not accurately segmented in the resultant segmentation, it will lead to at least one incorrect stenosis location. Furthermore, in the calculation of stenosis degree at small branches or stenotic locations, minor differences in cross-sectional area caused by segmentation can result in changes in the degree of stenosis. To illustrate this, consider the scenario where the cross-sectional areas $A_{\min}$ and $A_{\text{ref}}$ in the ground truth are 6 and 20, respectively. According to Equation (8), this would correspond to a severe stenosis point with a stenosis level of 0.7. Conversely, if the segmentation result calculates $A_{\min}$ and $A_{\text{ref}}$ as 7 and 21, respectively, the stenosis level becomes 0.667, which is classified as a moderate stenosis point. It is noteworthy that despite the minimal discrepancy in segmentation, this can result in a substantial variation in the degree of stenosis.

The efficacy of the stenosis detection method is contingent upon the quality of the segmentation results, with even minor inaccuracies potentially resulting in a reduction in the method's performance. Furthermore, the stenosis detection method relies exclusively on the vascular contour and morphological information for judgment. In actual clinical scenarios, additional information may be required for accurate assessment, including coronary artery plaque characteristics and Fractional Flow Reserve (FFR) measurements.

## 6 Conclusion

In this study, a 3D CTA coronary artery segmentation and detection framework is proposed, consisting of a novel MGFA model for segmentation, and a set of algorithms for automatic stenosis detection. In the segmentation model, the synergistic effects of several key modules, including myocardial region guidance, RFEE, and MSFF, enable the MGFA model to achieve remarkable results in coronary artery segmentation, significantly enhancing the accuracy and detail of the segmentation. Concurrently, the stenosis detection algorithm, through the close cooperation of modules such as centerline extraction, key point detection, and stenosis detection, provided a solid and precise basis for clinical decision-making in interventional treatment.The MGFA model has demonstrated exceptional performance in preclinical evaluations, exhibiting high practicality and reliability. It shows great potential for broad clinical application. The proposed segmentation and detection framework will provide valuable auxiliary suggestions for the precise diagnosis and effective treatment of coronary heart disease, thereby promoting the advancement of its diagnosis and treatment to new heights.

**Acknowledgments**

This study received support from the National Natural Science Foundation of China (Grant Numbers: 62476255, 82370513, 62303427, and 62106233), the Science and Technology Innovation Talent Project of Henan Province University (Grant Number 25HASTIT028), and the Henan Science and Technology Development Plan (Grant Number: 232102210010, 232102210062).